\documentclass[runningheads]{llncs}
\usepackage{graphicx}
\usepackage{amsmath,amssymb} 
\usepackage{color}
\makeatletter
\def\blfootnote{\xdef\@thefnmark{}\@footnotetext}
\makeatother

\begin{document}
\title{SpiderCNN: Deep Learning on Point Sets with Parameterized Convolutional Filters} 

\titlerunning{SpiderCNN}
%
\author{$\text{Yifan Xu}^{1 2*}$ \and
 $\text{Tianqi Fan}^{2*}$ \and
 Mingye Xu \inst{2} \and Long Zeng\inst{1} \and $\text{Yu Qiao}^{2 \dagger}$}
%
\authorrunning{Y. Xu, T. Fan, M. Xu, L. Zeng and Y. Qiao}
%

\institute{  Tsinghua University \\ \email{ \{xuyf16\}@mails.tsinghua.edu.cn}, \email{ \{zenglong\}@sz.tsinghua.edu.cn} \and
		Guangdong Key Lab of Computer Vision and Virtual Reality, \\
		SIAT-SenseTime Joint Lab, \\
		Shenzhen Institutes of Advanced Technology, Chinese Academy of Sciences\\
		\email{ \{tq.fan, my.xu, yu.qiao\}@siat.ac.cn}
	}
\maketitle              
\begin{abstract}
Deep neural networks have enjoyed remarkable success for various vision tasks, however it remains challenging to apply CNNs to domains lacking a regular underlying structures such as 3D point clouds. Towards this we propose a novel convolutional architecture, termed SpiderCNN, to efficiently extract geometric features from point clouds. SpiderCNN is comprised of  units called SpiderConv, which extend convolutional operations from regular grids to irregular point sets that can be embedded in $\mathbb{R}^n$, by parametrizing a family of convolutional filters. We design the filter as a product of a simple step function that captures local geodesic information and a Taylor polynomial that ensures the expressiveness. SpiderCNN inherits the multi-scale hierarchical architecture from classical CNNs, which allows it to extract semantic deep features. Experiments on ModelNet40\cite{chang2015shapenet} demonstrate that SpiderCNN achieves state-of-the-art accuracy $92.4 \%$ on standard benchmarks, and shows competitive performance on segmentation task.

\end{abstract}
\section{Introduction}

\blfootnote{$*$ These two authors contribute equally.~~~~~~~~~~~~~~~~~$\dagger$ Corresponding author.} 
\blfootnote{Work done during Yifan Xu's internship at SIAT.}

Convolutional neural networks are powerful tools for analyzing data that can naturally be represented as signals on regular grids, such as audio and images \cite{krizhevsky2012imagenet}. Thanks to the translation invariance of lattices in $\mathbb{R}^n$, the number of parameters in a convolutional layer is independent of the input size. Composing convolution layers and activation functions results in a multi-scale hierarchical learning pattern, which is shown to be very effective for learning deep representations in practice. 

With the recent proliferation of applications employing 3D depth sensors \cite{zhang2012microsoft} such as autonomous navigation, robotics and virtual reality, there is an increasing  demand for algorithms to efficiently analyze point clouds. 
However, point clouds are distributed irregularly in $\mathbb{R}^3$, lacking a canonical order and translation invariance, which prohibits using CNNs directly. One may circumvent this problem by converting point clouds to 3D voxels and apply 3D convolutions \cite{maturana2015voxnet}. However, volumetric methods are computationally inefficient because point clouds are sparse in 3D as they usually represent 2D surfaces. Although there are studies that improve the computational complexity, it may come with a performance trade off \cite{riegler2017octnet} \cite{brock2016generative}. Various studies are devoted to making convolution neural networks applicable for learning on non-Euclidean domains such as graphs or manifolds by trying to generalize the definition of convolution to functions on manifolds or graphs, enriching the emerging field of geometric deep learning \cite{bronstein2017geometric}. However, it is challenging theoretically because convolution cannot be naturally defined when the space does not carry a group action, and when the input data consists of different shapes or graphs, it is difficult to make a choice for convolutional filters.
\footnote{There is no canonical choice of a domain for these filters.}

\begin{figure}
\centering
\includegraphics[height=2.8cm]{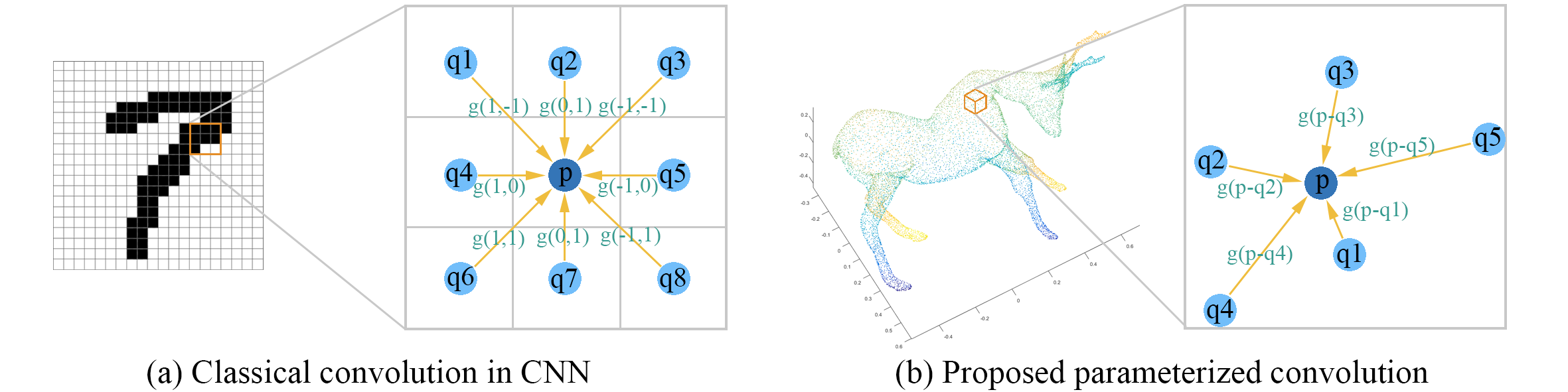}
\caption{ 
The integral formula for convolution between a signal $f$ and a filter $g$ is $ f~\ast~g (p) = \int_{q \in \mathbb{R}^n} f(q)g(p-q) dq$. Discretizing the integral formula on a set of points $P$ in  $\mathbb{R}^n$ gives $ f \ast g (p)  = \sum_{q \in P, \| p-q \| \leq r} f(q) g (p-q)$ if $g$ is supported in a ball of radius $r$.
{\bf (a)} when $P$ can be represented by regular grids, only $9$ values of a filter $g$ are needed to compute the convolution due to the translation invariance of the domain.
{\bf (b)} when the signal is on point clouds, we choose the filter $g$ from a parameterized family of function on $\mathbb{R}^3$.
}
\label{fig:spiderConv}
\end{figure}

In light of the above challenges, we propose an alternative convolutional architecture, SpiderCNN, which is designed to directly extract features from point clouds. We validate its effectiveness on classification and segmentation benchmarks.
By discretizing the integral formula of convolution as shown in Figure~\ref{fig:spiderConv}, and using a special family of parametrized non-linear functions on $\mathbb{R}^3$ as filters, we introduce a novel convolutional layer, SpiderConv, for point clouds.

The family of filters is designed to be expressive while still being feasible to optimize. We combine simple step functions, which are used to capture the coarse geometry described by local geodesic distance, with order-3 Taylor expansions, which ensure the filters are complex enough to capture intricate local geometric variations. Experiments in Section~\ref{experiments} show that  SpiderCNN with a relatively simple network architecture achieves the state-of-the-art performance for classification on ModelNet40 \cite{chang2015shapenet}, and shows competitive performance for segmentation on ShapeNet-Part \cite{chang2015shapenet}.

\section{Related Work}

First we discuss deep neural network based approaches that target point clouds data. Second, we give a partial overview of geometric deep learning. 

{\noindent \bf Point clouds as input:}
PointNet \cite{qi2017pointnet} is a pioneering work in using deep networks to directly process point sets. A spatial encoding of each point is learned through a shared MLP, and then all individual point features aggregate to a global signature through max-pooling, which is a symmetric operation that doesn't depend on the order of input point sequence. 

While PointNet works well to extract global features, its design limits its efficacy at encoding local structures. Various studies addressing this issue propose different grouping strategies of local features in order to mimic the hierarchical learning procedure at the core of classical convolutional neural networks.
PointNet++ \cite{qi2017pointnet++} uses iterative farthest point sampling to select centroids of local regions, and PointNet to learn the local pattern. Kd-Network \cite{klokov2017escape} subdivides the space using K-d trees, whose hierarchical structure serves as the instruction to aggregate local features at different scales. In SpiderCNN, no additional choice for grouping or sampling is needed, for our filters handle the issue automatically. 
 
The idea of using permutation-invariant functions for learning on unordered sets is further explored by DeepSet\cite{zaheer2017deep}. We note that the output of SpiderCNN does not depend on the input order by design.

{\noindent \bf Voxels as input:}
VoxNet \cite{maturana2015voxnet} and Voxception-ResNet \cite{brock2016generative} apply 3D convolution to a voxelization of point clouds. However, there is a high computational and memory cost associated with 3D convolutions. A variety of work \cite{riegler2017octnet}  \cite{engelcke2017vote3deep}  \cite{graham20173d} has aimed at exploiting the sparsity of voxelized point clouds to improve the computational and memory efficiency.
OctNet \cite{riegler2017octnet} modified and implemented convolution operations to suit a hybrid grid-octree data structure.
Vote3Deep \cite{engelcke2017vote3deep} uses a feature-centric voting scheme so that the computational cost is proportional to the number of points with non-zero features.
Sparse Submanifold CNN \cite{graham20173d} computes the convolution only at activated points whose number does not increase when the convolution layers are stacked. In comparison, SpiderCNN can use point clouds as input directly and can handle very sparse input. 

{\noindent \bf Convolution on non-Euclidean domain:}
There are two main philosophically different approaches to define convolutions for non-Euclidean domains: one is spatial and the other is spectral. The recent work ECC \cite{simonovsky2017dynamic} defines convolution-like operations on graphs where filter weights are conditioned on edge labels. Viewing point clouds as a graph, and taking the filters to be MLPs, SpiderCNN and ECC\cite{simonovsky2017dynamic} result in similar convolution. However, we show that our proposed family of filters outperforms MLPs.  

{\noindent \bf Spatial methods:}
GeodesicCNN \cite{masci2015geodesic} is an early attempt at applying neural networks to shape analysis. The philosophy behind GeodesicCNN is that for a Riemannian manifold, the exponential map identifies a local neighborhood of a point to a ball in the tangent space centered at the origin. The tangent plane is isomorphic to $\mathbb{R}^d$ where we know how to define convolution. 

Let $M$ be a mesh surface, and let $F: M \to \mathbb{R}$ be a function, GeodesicCNN first uses the patch operator $D$
to map a point $p$ and its neighbors $N(p)$ to the lattice $\mathbb{Z}^2 \subseteq \mathbb{R}^2$, and applies Equation~\ref{convolutionDis}. Explicitly, 
$F \ast g (p) = \sum_{j \in J} g_j (\sum_{q \in N(p)} w_j(u(p, q)) F(q))$,
where $u(p, q)$ represents the local polar coordinate system around $p$, $w_j(u)$ is a function to model the effect of the patch operator $D = \{D_j\}_{j \in J}$. By definition $D_j = \sum_{q \in N(p)} w_j(u(p, q)) F(q)$.
Later, AnisotrpicCNN \cite{boscaini2016learning} and MoNet \cite{monti2017geometric} further explore this framework by improving the choice for $u$ and $w_j$. MoNet \cite{monti2017geometric} can be understood as using mixtures of Gaussians as convolutional filters. We offer an alternative viewpoint. Instead of finding local parametrizations of the manifold, we view it as an embedded submanifold in $\mathbb{R}^n$ and design filters, which are more efficient for point clouds processing, in the ambient Euclidean space.

{\noindent \bf Spectral methods:}
We know that Fourier transform takes convolutions to multiplications. Explicitly, If $f, g: \mathbb{R}^n \to \mathbb{C}$, then $\widehat{f \ast g} = \hat{f} \cdot \hat{g}$. Therefore, formally we have $f \ast g =  {(\hat{f} \cdot \hat{g})}^{\vee}$,
\footnote{If $h$ is a function, then $\hat{h}$ is the Fourier transform, and $h^{\vee}$ is its inverse Fourier transform.}
which can be used as a definition for convolution on non-Euclidean domains where we know how to take Fourier transform. 

Although we do not have Fourier theory on a general space without any equivariant structure, on Riemannian manifolds or graphs there are generalized notions of Laplacian operator. Taking Fourier transform in $\mathbb{R}^n$ could be formally viewed as finding the coefficients in the expansion of the eigenfunctions of the Laplacian operator. To be more precise, recall that 
\begin{equation}
\hat{f} (\xi) = \int_{\mathbb{R}^n} f(x) \exp{(- 2 \pi i x \cdot \xi)} d \xi,
\end{equation}
and  $\{ \exp{(- 2 \pi i x \cdot \xi}) \}_{\xi \in \mathbb{R}^n}$ are eigen-functions for the Laplacian operator 
$\Delta = \sum_{i = 1}^n \frac{\partial}{\partial x_i}$.
Therefore, if $U$ is the matrix whose columns are eigenvectors of the graph Laplacian matrix and $\Lambda$ is the vector of corresponding eigenvalues, for $F, g$ two functions on the vertices of the graph, then $F \ast g  = U (U^{T}F \odot U^{T}g)$, where $U^{T}$ is the transpose of $U$ and $\odot$ is the Hadamard product of two matrices.
Since being compactly supported in the spatial domain translates into being smooth in the spectral domain, it is natural to choose $U^Tg$ to be smooth functions in $\Lambda$. For instance, ChebNet \cite{defferrard2016convolutional} uses Chebyshev polynomials that reduces the complexity of filtering, and CayleyNet \cite{levie2017cayleynets} uses Cayley polynomials which allows efficient computations for localized filters in restricted frequency bands of interest. 

When analyzing different graphs or shapes, spectral methods lack abstract motivations, because different spectral domains cannot be canonically identified. SyncSpecCNN \cite{yi2017syncspeccnn} proposes a weight sharing scheme to align spectral domains using functional maps. Viewing point clouds as data embedded in $\mathbb{R}^3$, SpiderCNN can learn representations that are robust to spatial rigid transformations with the aid of data augmentation.

\section{SpiderConv}

In this section, we describe SpiderConv, which is the fundamental building block for SpiderCNN. First, we discuss how to define a convolutional layer in neural network when the inputs are features on point sets in $\mathbb{R}^n$. Next we introduce a special family of convolutional filters. Finally, we give details for the implementation of SpiderConv with multiple channels and the approximations used for computational speedup. 

\subsection{Convolution on point sets in $\mathbb{R}^n$}

An image is a function on regular grids $F: \mathbb{Z}^2 \to \mathbb{R}$. Let $W$ be a $(2m+1)\times (2m+1)$ filter matrix, where $m$ is a positive integer, the convolution in classical CNNs is 
\begin{equation} \label{convolutionDis}
F \ast W (i, j) = \sum_{s = -m}^m \sum_{t = -m}^{m} F(i - s, j - t)W(s, t),
\end{equation}
which is the discretization of the following integration
\begin{equation} \label{convolutionMath}
f \ast g (p) = \int_{\mathbb{R}^2} f(q) g (p - q) d q,
\end{equation}
if $f, g: \mathbb{R}^2 \to \mathbb{R}$, such that $f(i, j) = F(i, j)$ for $(i, j) \in \mathbb{Z}^2$ and $g(s,t) = W(s, t)$ for $s, t \in \{-m,-m+1, ... , m-1, m \} $ and $g$ is supported in $[-m, m] \times [-m, m]$. 

Now suppose that $F$ is a function on a set of points $P$ in $\mathbb{R}^n$. Let $g: \mathbb{R}^n \to \mathbb{R}$ be a filter supported in a ball centered at the origin of radius $r$. It is natural to define SpiderConv with input $F$ and filter $g$ to be the following:
\begin{equation} \label{SpiderCNN}
F \ast g (p) = \sum_{q \in P, \| q - p\| \leq r}  F(q) g(p-q).
\end{equation}
Note that when $P = \mathbb{Z}^2$ is a regular grid, Equation~\ref{SpiderCNN} reduces to Equation~\ref{convolutionMath}.  Thus the classical convolution can be seen as a special case of SpiderConv. Please see Figure~\ref{fig:spiderConv} for an intuitive illustration.

\begin{figure}
\centering
\includegraphics[height=6.5cm]{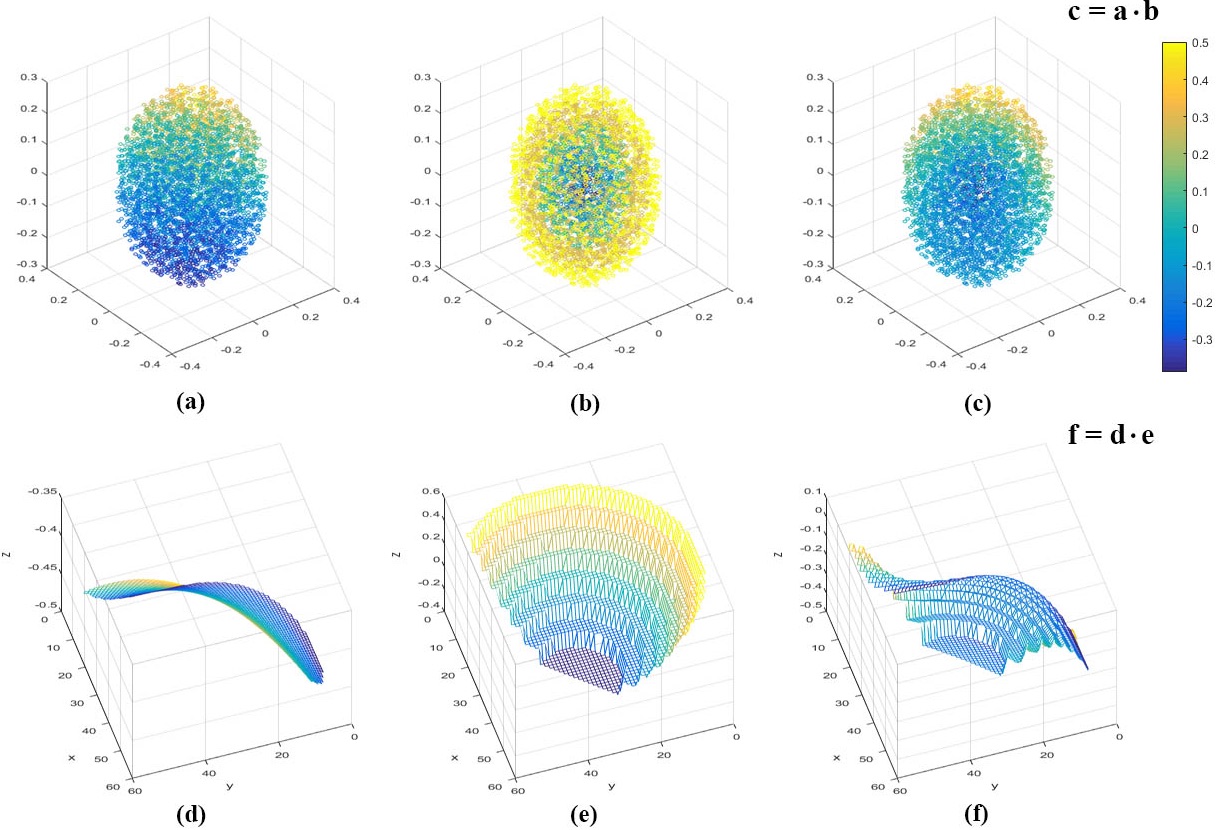} \label{fig13}
\caption{ 
Visualization of a filter in the family $\{ g_w\}$. {\bf (a)} is the scatter plot (color represents the value of the function) of $g^{Taylor}(x, y, z) = 1 + x + y + z + xy + xz + yz + xyz$. {\bf{(b)}}  is the scatter plot of $g^{step}(x, y, z) = \frac{i+1}{8}$ if $\frac{i}{8} \leq \sqrt{x^2 + y^2 + z^2} < \frac{i+1}{8}$, when $i = 0, 1, ... , 7$. {\bf (c)} is the scatter plot of the product $g =g^{Taylor} \cdot g^{step}$. In the second row, {\bf (d)} {\bf (e)} {\bf (f)} are  the graphs of $g^{Taylor}$ , $g^{step}$   and $g$  respectively when restricting their domain to the plane $z = 0$ (the $Z$-axis represents the value of the function).
}
\label{fig:filters}
\end{figure}

In SpiderConv, the filters are chosen from a parametrized family $\{ g_w \}$ (See Figure~\ref{fig13} for a concrete example) which is piece-wise differentiable in $w$. During the training of SpiderCNN, the parameters $w \in \mathbb{R}^d$ are optimized through SGD algorithm, and the gradients are computed through the formula 
$\frac{\partial}{\partial w_i} F \ast g_w (p) = \sum_{q \in P, \| q - p\| \leq r}  F(q) \frac{\partial}{\partial w_i}  g_w(p-q)$, 
where $w_i$ is the $i$-th component of $w$.

\subsection{A special family of filters $\{ g_w \}$ } \label{TS}

A natural choice is to take $g_w$ to be a multilayer perceptron (MLP) network, because theoretically an MLP with one hidden layer can approximate an arbitrary continuous function \cite{hornik1991approximation}. However, in practice we find that MLPs do not work well. 
One possible reason is that MLP fails to account for the geometric prior of 3D point clouds. Another possible reason is that to ensure sufficient expressiveness the number of parameters in a MLP needs to be sufficiently large, which makes the optimization problem difficult.

To address the above issues, we propose the following family of filters $\{ g_w \}$:
\begin{equation}
g_{w}(x, y, z) = g^{Step}_{w^S} (x, y, z) \cdot g^{Taylor}_{w^T} (x, y, z),
\end{equation}
with $w = (w^S, w^T)$ is the concatenation of two vectors $w^S = (w^S_i)$ and $w^T = (w^T_i)$,
\footnote{Here we use the notation $v = (v_i)$ to represent that $v_i \in \mathbb{R}$ is the $i$-th component of the vector $v$.}
where
\begin{equation} \label{filterStep}
g^{Step}_{w^S} (x, y, z) = w^S_i \text{ if } r_i \leq \sqrt{x^2 + y^2 + z^2}  < r_{i +1},
\end{equation}
with $r_0 = 0 < r_1 < r_2 ... < r_N$,
and
\begin{align} \label{Taylor}
\begin{split}
g^{Taylor}_{w^T} (x, y, z) &= w^T_0 + w^T_1 x + w^T_2 y + w^T_3 z  + w^T_4 xy + w^T_5 yz + w^T_6 xz + w^T_7 x^2 \\
				&+ w^T_8 y^2 + w^T_9 z^2  + w^T_{10} xy^2 + w^T_{11} x^2y + w^T_{12} y^2z + w^T_{13} yz^2 \\
				&+ w^T_{14} x^2z + w^T_{15} xz^2 + w^T_{16} xyz + w^T_{17} x^3 + w^T_{18} y^3 + w^T_{19} z^3.
\end{split}
\end{align}

The first component $g^{Step}_{w^S}$ is a step function in the radius variable of the local polar coordinates around a point. It encodes the local geodesic information, which is a critical quantity to describe the coarse local shape. Moreover, step functions are relatively easy to optimize using SGD.

The order-3 Taylor term $g^{Taylor}_{w^T}$ further enriches the complexity of the filters, complementary to $g^{Step}_{w^S}$ since it also captures the variations of the angular component. Let us be more precise about the reason for choosing Taylor expansions here from the perspective of interpolation.
We can think of the classical 2D convolutional filters as a family of functions interpolating given values at 9 points $\{ (i, j) \}_{i, j \in \{ -1, 0, 1\} }$, and the $9$ values serve as the parametrization of such a family. Analogously, in 3D consider the vertices of a cube $\{ (i, j, k) \}_{i, j, k = 0, 1}$, assume that at the vertex $(i, j, k)$ the value $a_{i, j, k}$ is assigned. The trilinear interpolation algorithm gives us a function of the form 
\begin{equation} \label{trilinear}
f_{w^T}(x, y, z) = w^T_0 + w^T_1 x + w^T_2 y + w^T_3 z + w^T_4 xy + w^T_5 yz + w^T_6 xz + w^T_{16} xyz,
\end{equation} 
where $w^T_i$'s are linear functions in $c_{ijk}$. Therefore $f_{w^T}$ is a special form of $g^{Taylor}_{w^T}$, and by varying $w^T$, the family $\{ g^{Taylor}_{w^T}\}$ can interpolate arbitrary values at the vertexes of a cube and capture rich spatial information.

\subsection{Implementation} \label{implementation}
The following approximations are used based on the uniform sampling process constructing the point clouds:
\begin{enumerate}
\item 
K-nearest neighbors are used to measure the locality instead of the radius, so the summation in Equation~\ref{SpiderCNN} is over the K-nearest neighbors of $p$.  
\item
The step function $g^{Step}_{w^T}$ is approximated by a permutation. Explicitly, let $X$ be the $1 \times K$ matrix indexed by the K-nearest neighbors of $p$ including $p$, and $X(1, i)$ is a feature at the $i$-th K-nearest neighbors of $p$. Then $F \ast g^{Step}_{w^T} (p)$ is approximated by $Xw$, where $w$ is a $K \times 1$ matrix with $w(i, 1)$ corresponds to $w^T_i$  in Equation~\ref{filterStep}. 
\end{enumerate}

Later in the article, we omit the parameters $w$, $w^S$ and $w^T$, and just write $g = g^{Step} \cdot g^{Taylor}$ to simplify our notations.

The input to SpiderConv is a $c_1$-dimensional feature on a point cloud $P$, and is represented as $F = (F_1, F_2, ... , F_{c_1})$ where $F_v : P \to \mathbb{R}$. The output of a SpiderConv is a $c_2$-dimensional feature on the point cloud $\tilde{F} = (\tilde{F}_1, \tilde{F}_2, ... , \tilde{F}_{c_2})$ where $\tilde{F}_i : P \to \mathbb{R}$. 
Let $p$ be a point in the point cloud, and $q_1, q_2, ... , q_K$ are its $K$-nearest neighbors in order. Assume $g^{Step}_{i, v, t} (p - q_j) = w^{(i, v, t)}_{j}$, where $t = 1, 2, ... , b$ and $v = 1, 2, ... , c_1$ and $i = 1, 2, ... c_2$. Then a SpiderConv with $c_1$ in-channels, $c_2$ out-channels and $b$ Taylor terms is defined via the formula:
$\tilde{F}_i(p) =\sum_{v = 1}^{c_1} \sum_{j = 1}^K g_i(p-q_j)F_{v}(q_j)$,
where 
$g_i(p - q_j) = \sum_{t = 1}^b g_t^{Taylor} (p - q_j) w_{j}^{(i, v, t)}$,
and $g_t^{Taylor}$ is in the parameterized family $\{ g^{Taylor}_{w^T} \}$ for $t = 1, 2, ..., b$.

\section{Experiments} \label{experiments}
We analyze and evaluate SpiderCNN on 3D point clouds classification and segmentation. We empirically examine the key hyper-parameters of a 3-layer SpiderCNN, and compare our models with the state-of-the-art methods.

{\noindent \bf Implementation Details:} All models are prototyped with Tensorflow 1.3 on 1080Ti GPU and trained using the Adam optimizer with a learning rate of $10^{-3}$. A dropout rate of 0.5 is used with the the fully connected layers. Batch normalization is used at the end of each SpiderConv with decay set to 0.5. 
On a GTX 1080Ti, the forward-pass time of a SpiderConv layer (batch size 8) with in-channel 64 and out-channel 64 is 7.50 ms. For the 4-layer SpiderCNN (batch size 8), the total forward-pass time is 71.68 ms. 

\subsection{Classification on ModelNet40} \label{modelnet}

ModelNet40 \cite{chang2015shapenet} contains 12,311 CAD models which belong to 40 different categories with 9,843 used for training and 2,468 for testing. We use the source code for PointNet\cite{qi2017pointnet} to sample 1,024 points uniformly and compute the normal vectors from the mesh models. The same data augmentation strategy as \cite{qi2017pointnet} is applied: the point cloud is randomly rotated along the up-axis and the position of each point is jittered by a Gaussian noise with zero mean and 0.02 standard deviation. The batch size is 32 for all the experiments in Section~\ref{modelnet}. We use the $(x, y, z)$-coordinates and normal vectors of the 1,024 points as the input for SpiderCNN for the experiments on ModelNet40 unless otherwise specified. 

\begin{figure}
\centering
\includegraphics[height=3.2cm]{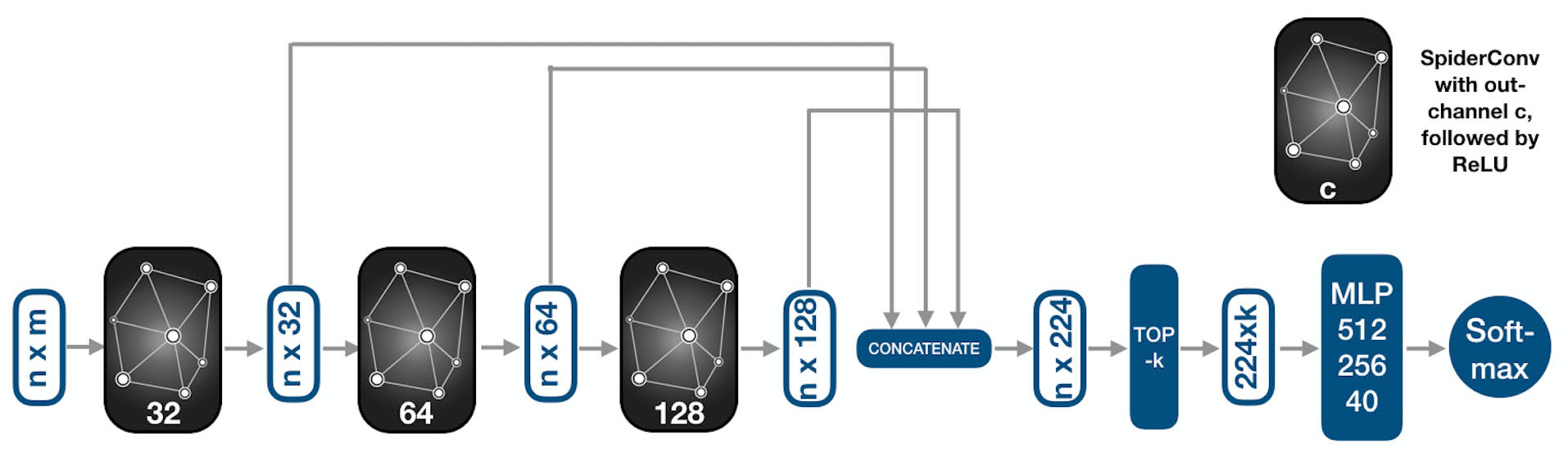} 
\caption{The architecture of a 3-layer SpiderCNN used in ModelNet40 classification.}
\label{fig:spidercnn}
\end{figure}

{\noindent \bf 3-layer SpiderCNN: }
Figure~\ref{fig:spidercnn} illustrates a SpiderCNN with 3 layers of SpiderConvs each with 3 Taylor terms, and the respective out-channels for each layer being 32, 64, 128. 
\footnote{See Section~\ref{implementation} for the definition of a SpiderConv with $c_1$ in-channels, $c_2$ out-channels and $b$ Taylor terms.} 
ReLU activation function is used here. The output features of the three SpiderConvs are concatenated in the end. Top-$k$ pooling among all the points is used to extract global features. 

\begin{figure}
\centering
\includegraphics[height=3.5cm]{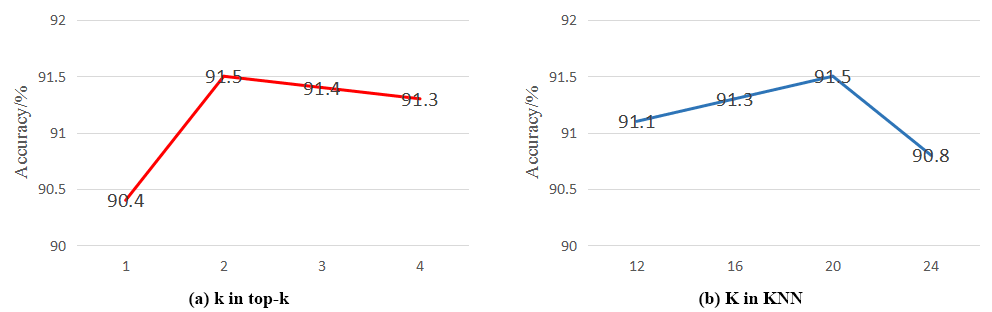} 
\caption{
{On ModelNet40 \bf (a)} shows the effect of number of pooled features on the accuracy of 3-layer SpiderCNN with 20-nearest neighbors. 
{\bf (b)} shows the effect of nearest neighbors in SpiderConv on the accuracy of 3-layer SpiderCNN with top-2 pooling . 
}
\label{fig:spiderCNNhyper}
\end{figure}

\begin{figure}
\centering
\includegraphics[height=4.7cm]{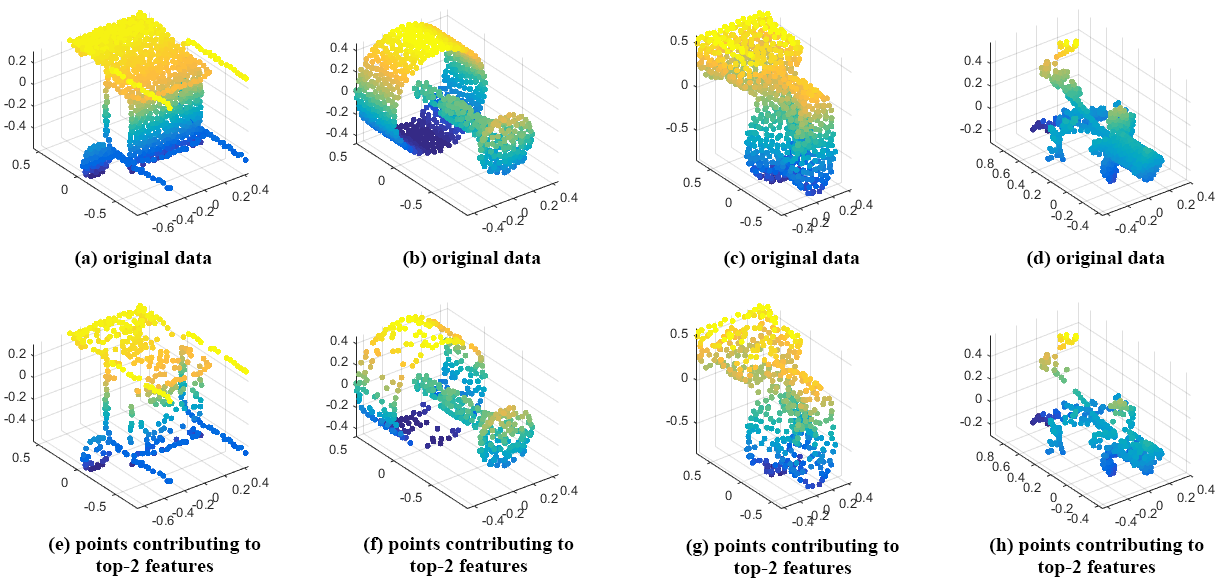} 
\caption{Visualization of the effect of top-$k$ pooling. Edge points and points with non-zero curvature are preserved after pooling.
{\bf (a) (b) (c) (d)} are the original input point clouds. 
{\bf (e) (f) (g) (h)} are points contributing to features extracted via top-2 pooling.
}
\label{fig:topk1}
\end{figure}

Two important hyperparameters in SpiderCNN are studied: the number of nearest neighbors $K$ chosen in SpiderConv, and the number of pooled features $k$ after the concatenation. The results are summarized in Figure~\ref{fig:spiderCNNhyper}. The number of nearest-neighbors $K$ is analogous to size of the filter in the usual convolution. We see that 20 is the optimal choice among 12, 16, 20, and 24-nearest neighbors.
In Figure~\ref{fig:topk1} we provide visualization for top-2 pooling. The points that contribute to the top-2 pooling features are plotted. We see that similar to PointNet, Spider~CNN picks up representative critical points. 

{\noindent \bf SpiderCNN + PointNet: }
We train a 3-layer SpiderCNN (top-2 pooling and 20-nearest neighbors) and PointNet with only $(x,y,z)$-coordinates as input to predict the classical robust local geometric descriptor FPFH \cite{rusu2009fast} on point clouds in ModelNet40. The training loss of SpiderCNN is only $\frac{1}{4}$ that of PointNet's.
As a result, we believe that a 3-layer SpiderCNN and PointNet are complementary to each other, for SpiderCNN is good at learning local geometric features and PointNet is good at capturing global features. By concatenating the 128 dimensional features from PointNet with the 128 dimensional features from SpiderCNN, we improve the classification accuracy to $92.2 \%$. 

{\noindent \bf 4-layer SpiderCNN: } 
Experiments show that 1-layer SpiderCNN with a SpiderConv of 32 channels can achieve classification accuracy $85.5\%$, and the performance of SpiderCNN improves with the increasing number of layers of SpiderConv. A 4-layer SpiderCNN consists of SpiderConv with out-channels 32, 64, 128, and 258. Feature concatenation, 20-nearest neighbors and top-2 pooling are used. To prevent overfitting, while training we apply the data augmentation method DP (random input dropout) introduced in \cite{qi2017pointnet++}. 
Table~\ref{table:modelnet40} shows a comparison between SpiderCNN and other models. The 4-layer SpiderCNN achieves accuracy of $92.4\%$ which improves over the best reported result of models with input 1024 points and normals.
For 5 runs, the mean accuracy of a 4-layer SpiderCNN is $92.0 \%$.

\setlength{\tabcolsep}{4pt}
\begin{table}
\begin{center}
\caption{Classification accuracy of SpiderCNN and other models on ModelNet40.} 
\label{table:modelnet40}
\begin{tabular}{lll}
\hline\noalign{\smallskip}
Method & Input & Accuracy\\
\noalign{\smallskip}
\hline
\noalign{\smallskip}
Subvolume \cite{qi2016volumetric} & voxels & 89.2
\\
VRN Single \cite{brock2016generative} & voxels & 91.3
\\
OctNet \cite{riegler2017octnet} & hybrid grid octree & 86.5
\\
ECC \cite{simonovsky2017dynamic} & graphs & 87.4
\\
\hline
Kd-Network \cite{klokov2017escape} (depth 15) & 1024 points & 91.8
\\
PointNet \cite{qi2017pointnet}& 1024 points & 89.2
\\
PointNet++ \cite{qi2017pointnet++}& 5000 points+normal & 91.9
\\
\hline
SpiderCNN + PointNet & 1024 points+normal & 92.2
\\
SpiderCNN (4-layer) & 1024 points+normal & \bf{92.4}
\\ 
\hline
\end{tabular}
\end{center}
\end{table}
\setlength{\tabcolsep}{1.4pt}

\begin{figure}
\centering
\includegraphics[height=2.2cm]{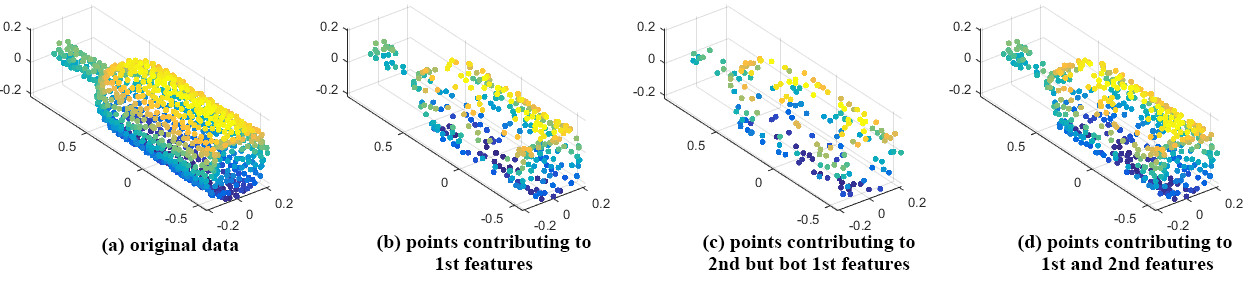} 
\caption{
Top-2 pooling learns rich features and fine geometric details.
}
\label{fig:topk2}
\end{figure}

{\noindent \bf Ablative Study:}
Compared to max-pooling, top-2 pooling enables the model to learn richer geometric information. For example, in Figure~\ref{fig:topk2}, we see top-2 pooling preserves more points where the curvature is non-zero. Using max-pooling, the classification accuracy is $92.0\%$ for a 4-layer SpiderCNN, and is $90.4\%$ for a 3-layer SpiderCNN. 
In comparison, using top-2 pooling, the accuracy is $92.4\%$ for a 4-layer SpiderCNN, and  is $91.5\%$ for a 3-layer SpiderCNN.

MLP filters do not perform as well in our setting. 
The accuracy of a 3-layer SpiderCNN is $71.3 \%$ with $g_w = \text{MLP}(16, 1)$, and is $72.8 \%$ with $g_w = \text{MLP}(16, 32, 1)$.

Without normals, the accuracy of a 4-layer SpiderCNN using only the 1,024 points is $90.5\%$. Using normals extracted from the 1,024 input points via orthogonal distance regression, the accuracy of a 4-layer SpiderCNN is $91.8\%$.

\begin{figure}
\centering
\includegraphics[height=3cm]{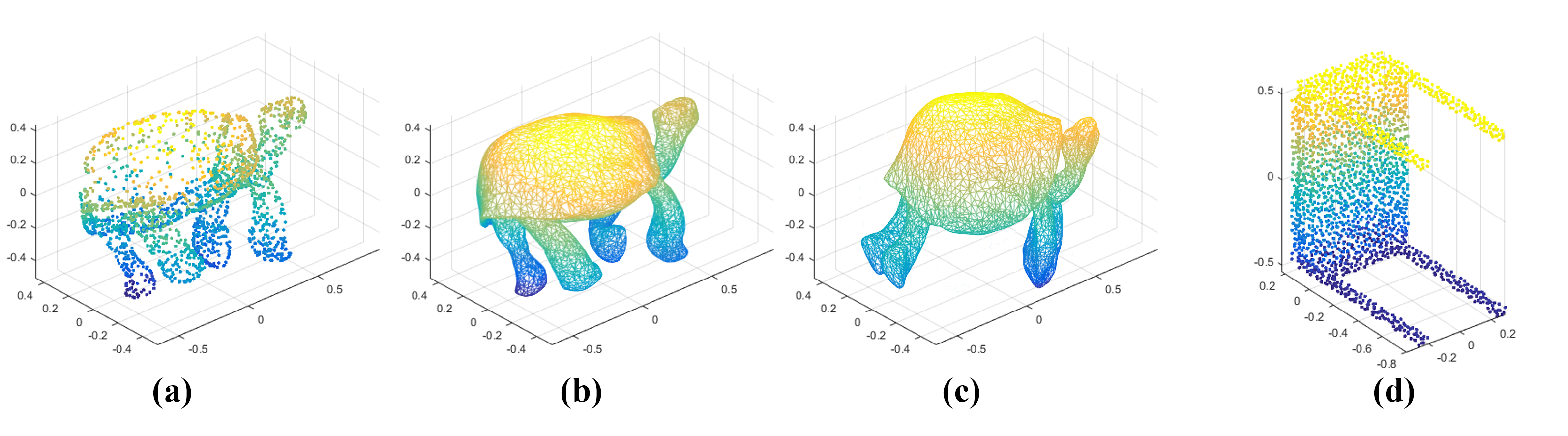} 
\caption{
{\bf (b)} and { \bf(c) } are shapes in SHREC15. {\bf (d) } is a shape in ModelNet40.
{\bf (a)} is the point cloud sampled from {\bf (b)}.
}
\label{fig:shrec15}
\end{figure}

\subsection{Classification on SHREC15}

SHREC15 is a dataset for non-rigid 3D shape retrieval. It consists of 1,200 watertight triangle meshes divided in 50 categories. On average 10,000 vertices are stored in one mesh model. Comparing to ModelNet40, SHREC15 contains more complicated local geometry and non-rigid deformation of one object. See Figure~\ref{fig:shrec15} for a comparison. 
1,192 meshes are used with 895 for training and 297 for testing. 
\setlength{\tabcolsep}{4pt}
\begin{table}
\begin{center}
\caption{Classification accuracy on SHEREC15.}
\label{table:shrec15}
\begin{tabular}{lllllll}
\hline\noalign{\smallskip}
Method & Input & Accuracy
\\
\noalign{\smallskip}
\hline
\noalign{\smallskip}
SVM + HKS & features & 56.9
\\
SVM + WKS& features & 87.5
\\
SVM + FPFH & features & 80.8
\\
PointNet & points & 69.4
\\
PointNet++ \cite{qi2017pointnet++} & points & 60.2
\\
PointNet++(our implementation) & points & 94.1
\\
SpiderCNN (4-layer) & points  & {\bf 95.8}
\\
\hline
\end{tabular}
\end{center}
\end{table}
\setlength{\tabcolsep}{1.4pt}
We compute three intrinsic shape descriptors (Heat Kernel Signature, Wave Kernel Signature and Fast Point Feature Histograms) for deformable shape analysis from the mesh models. 1,024 points are sampled uniformly randomly from the vertices of a mesh model, and the $(x, y, z)$-coordinates are  used as the input for SpiderCNN, PointNet and PointNet++. 
We use SVM with linear kernel when the inputs are classical shape descriptors. Table~\ref{table:shrec15} summarizes the results. We see that SpiderCNN outperforms the other methods.

\subsection{Segmentation on ShapeNet Parts}

ShapeNet Parts consists of 16,880 models from 16 shape categories and 50 different parts in total, with a 14,006 training and 2,874 testing split. Each part is annotated with 2 to 6 parts. The mIoU is used as the evaluation metric, computed by taking the average of all part classes.
\begin{figure}
\centering
\includegraphics[height=3cm]{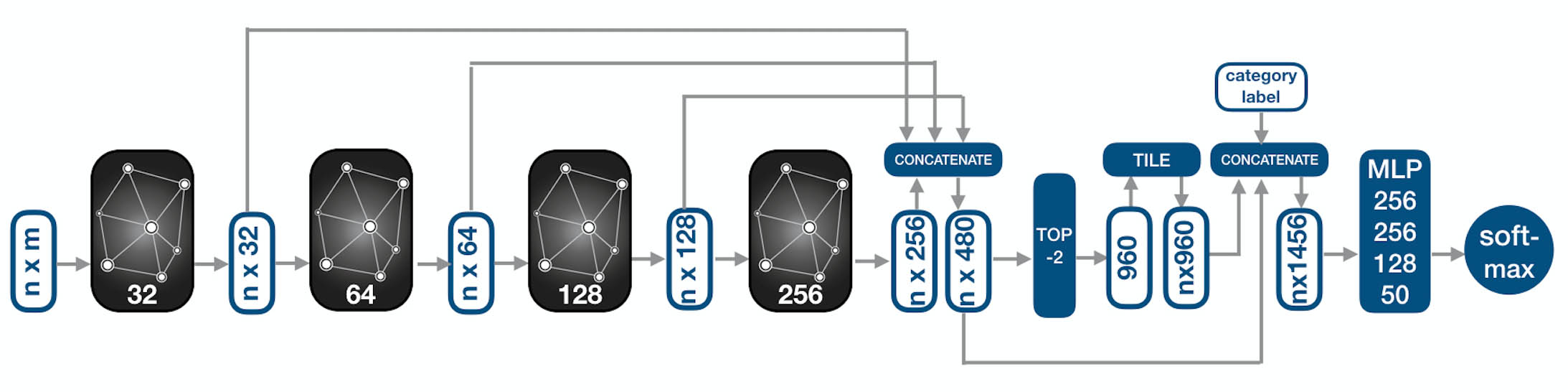} 
\caption{The SpiderCNN architecture used in the ShapeNet Part segmentation task.}
\label{fig:pargsegarch}
\end{figure}
A 4-layer SpiderCNN whose architecture is shown in Figure~\ref{fig:pargsegarch} is trained with batch of 16. We use points with their normal vectors as the input and assume that the category labels are known. The results are summarized in Table~\ref{table:scanseg}. For 4 runs, the mean of mean IoU of SpiderCNN is $85.24$. We see that SpiderCNN achieves competitive results despite a relatively simple network architecture.

\begin{table}
\begin{center}
\caption{Segmentation results on ShapeNet Part dataset. Mean IoU and IoU for each categories are reported.}
\label{table:scanseg}
\begin{tabular}{l | l | llllllllllllllll}
\hline\noalign{\smallskip}
   & {\tiny mean}  & {\tiny aero}  & {\tiny bag} & {\tiny cap} & {\tiny car} & {\tiny chair} & {\tiny ear} & {\tiny guitar} & {\tiny knife} & {\tiny lamp} & {\tiny laptop} & {\tiny motor} & {\tiny mug} & {\tiny pistol} & {\tiny rocket} & {\tiny skate}  & {\tiny table}   
\\
  &         &  &      &         &        &  &    {\tiny ph}         &         &         &             &  &         &            &            & {\tiny board} &       
 \\
\noalign{\smallskip}
\hline
\noalign{\smallskip}
{\tiny PN \cite{qi2017pointnet}}  & {\tiny 83.7}  & {\tiny 83.4}  & {\tiny 78.7} & {\tiny 82.5} & {\tiny 74.9} & {\tiny 89.6} & {\tiny 73.0} & {\tiny 91.5} & {\tiny 85.9} & {\tiny 80.8} & {\tiny 95.3} & {\tiny65.2} & {\tiny 93.0} & {\tiny 81.2} & {\tiny 57.9} & {\tiny 72.8}  & {\tiny 80.6}    
\\
{\tiny PN++\cite{qi2017pointnet++}}& {\tiny 85.1}  & {\tiny 82.4}  & {\tiny 79.0} & {\bf \tiny 87.7} & {\tiny 77.3} & {\bf \tiny 90.8} & {\tiny 71.8} & {\tiny 91.0} & {\tiny 85.9} & {\tiny 83.7} & {\tiny 95.3} & {\bf \tiny 71.6} & {\bf \tiny 94.1} & {\tiny 81.3} & {\tiny 58.7} & {\tiny 76.4}  & {\tiny 82.6}  
\\
{\tiny Kd-Net \cite{klokov2017escape}}  & {\tiny 82.3}  & {\tiny80.1}  & {\tiny 74.6} & {\tiny 74.3} & {\tiny 70.3} & {\tiny 88.6} & {\tiny 73.5} & {\tiny 90.2} & {\tiny 87.2} & {\tiny 81.0} & {\tiny94.9} & {\tiny 57.4} & {\tiny 86.7} & {\tiny 78.1} & {\tiny 51.8} & {\tiny 69.9}  & {\tiny 80.3}       
\\
{\tiny SSCNN \cite{yi2017syncspeccnn}}  & {\tiny 84.7}  & {\tiny 81.6}  & {\bf \tiny 81.7} & {\tiny 81.9} & {\tiny 75.2} & {\tiny 90.2} & {\tiny 74.9} & {\bf \tiny 93.0} & {\tiny 86.1} & {\bf \tiny 84.7} & {\tiny 95.6} & {\tiny 66.7} & {\tiny 92.7} & {\tiny 81.6} & {\bf \tiny 60.6} & {\bf \tiny 82.9}  & {\tiny 82.1}      
\\
\hline
{\tiny SpiderCNN}  & {\bf \tiny 85.3}  & {\bf \tiny83.5}  & {\tiny 81.0} & {\tiny 87.2} & {\bf \tiny 77.5} & {\tiny 90.7} & {\bf \tiny 76.8} & {\tiny 91.1} & {\bf \tiny 87.3} & {\tiny 83.3} & {\bf \tiny 95.8} & {\tiny 70.2} & {\tiny 93.5} & {\bf \tiny 82.7} & {\tiny 59.7} & {\tiny 75.8}  & {\bf \tiny82.8}     
\\
\hline
\end{tabular}
\end{center}
\end{table}

\begin{figure}
\centering
\includegraphics[height=2.1cm]{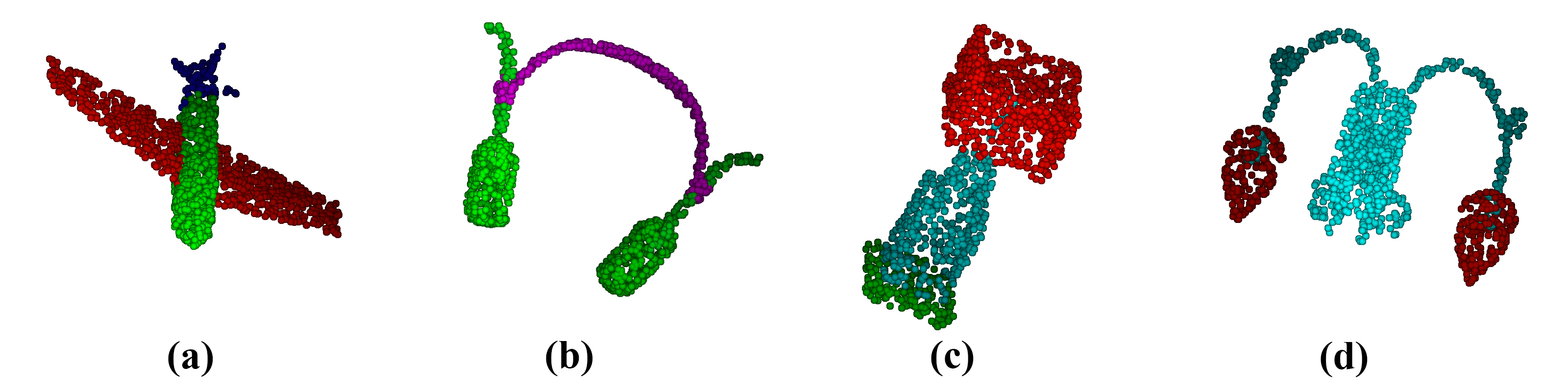} 
\caption{Some examples of the segmentation results of SpiderCNN on ShapeNet Part.}
\label{fig:sipidercnn}
\end{figure}

\section{Analysis}

In this section, we conduct additional analysis and evaluations on the robustness of SpiderCNN, and provide visualization for some of the typical  learned filters from the first layer of SpiderCNN.

\begin{figure}
\centering
\includegraphics[height=3.6cm]{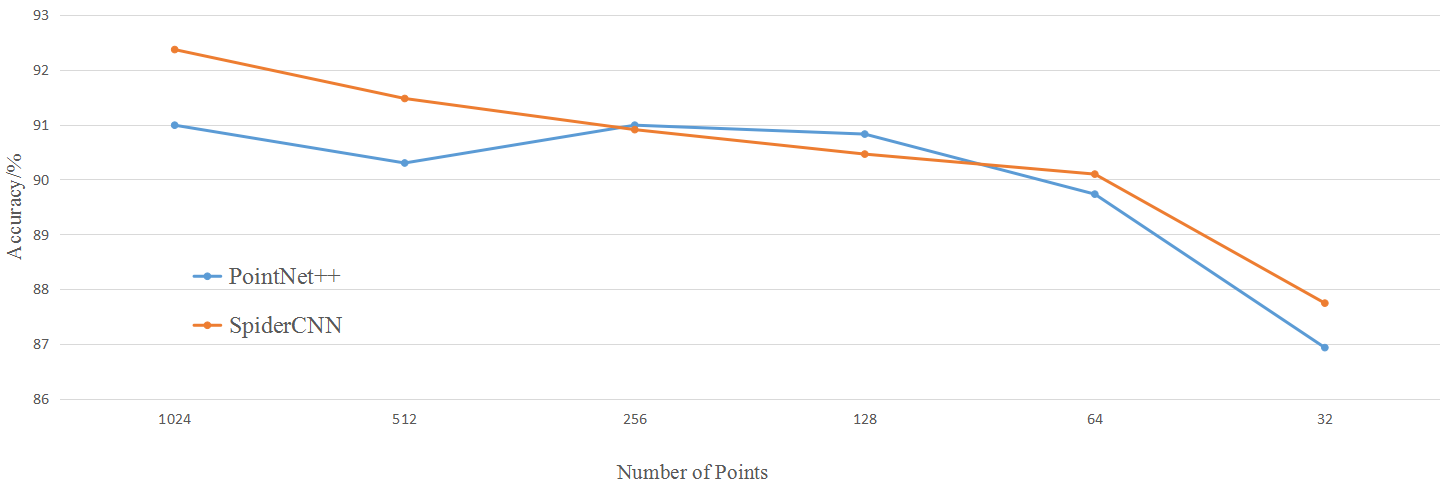} 
\caption{
Classification accuracy of SpiderCNN and PointNet++ with different number of input points on ModelNet40.
}
\label{fig:robust}
\end{figure}

{\noindent \bf Robustness:}
We study the effect of missing points on SpiderCNN. Following the setting for experiments in Section~\ref{modelnet}, we train a 4-layer SpiderCNN and PointNet++ with 512, 248, 128, 64 and 32 points and their normals as input. The results are summarized in Figure~\ref{fig:robust}. We see that even with only 32 points, SpiderCNN obtains $87.7 \%$ accuracy.

\begin{figure} 
\centering 
\includegraphics[height=3cm]{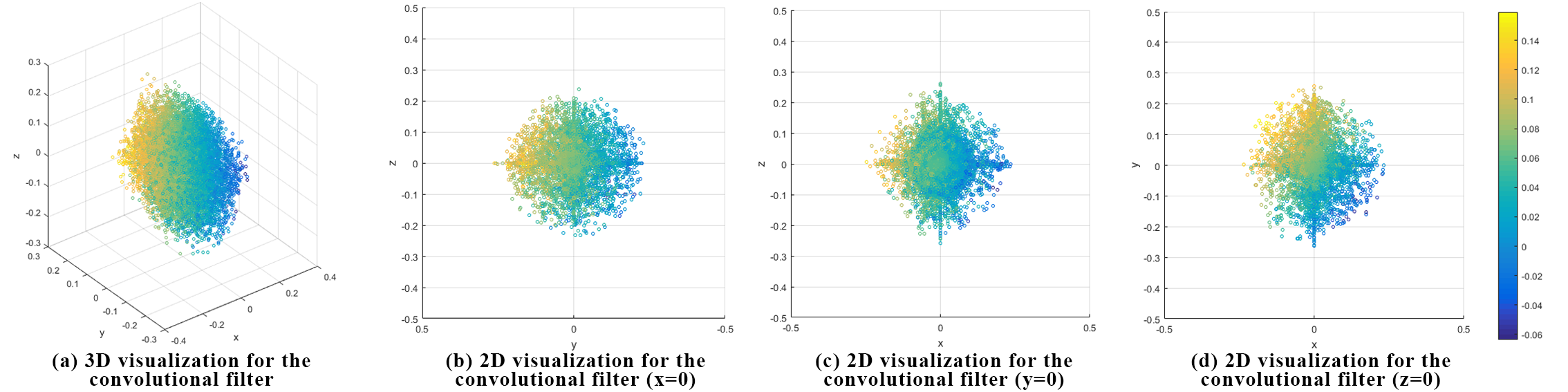} 
\caption{
Visualization of  for the convolutional filters learned in in the first layer of SpiderCNN. 
}
\label{fig:3dvisual}
\end{figure}

{\noindent \bf Visualization:}
In Figure~\ref{fig:3dvisual}, we scatter plot the convolutional filters $g_w(x, y, z)$ learned in the first layer of SpiderCNN and the color of a point represents the value of $g_w$ at the point.

\begin{figure}
\centering
\includegraphics[height=6.5cm]{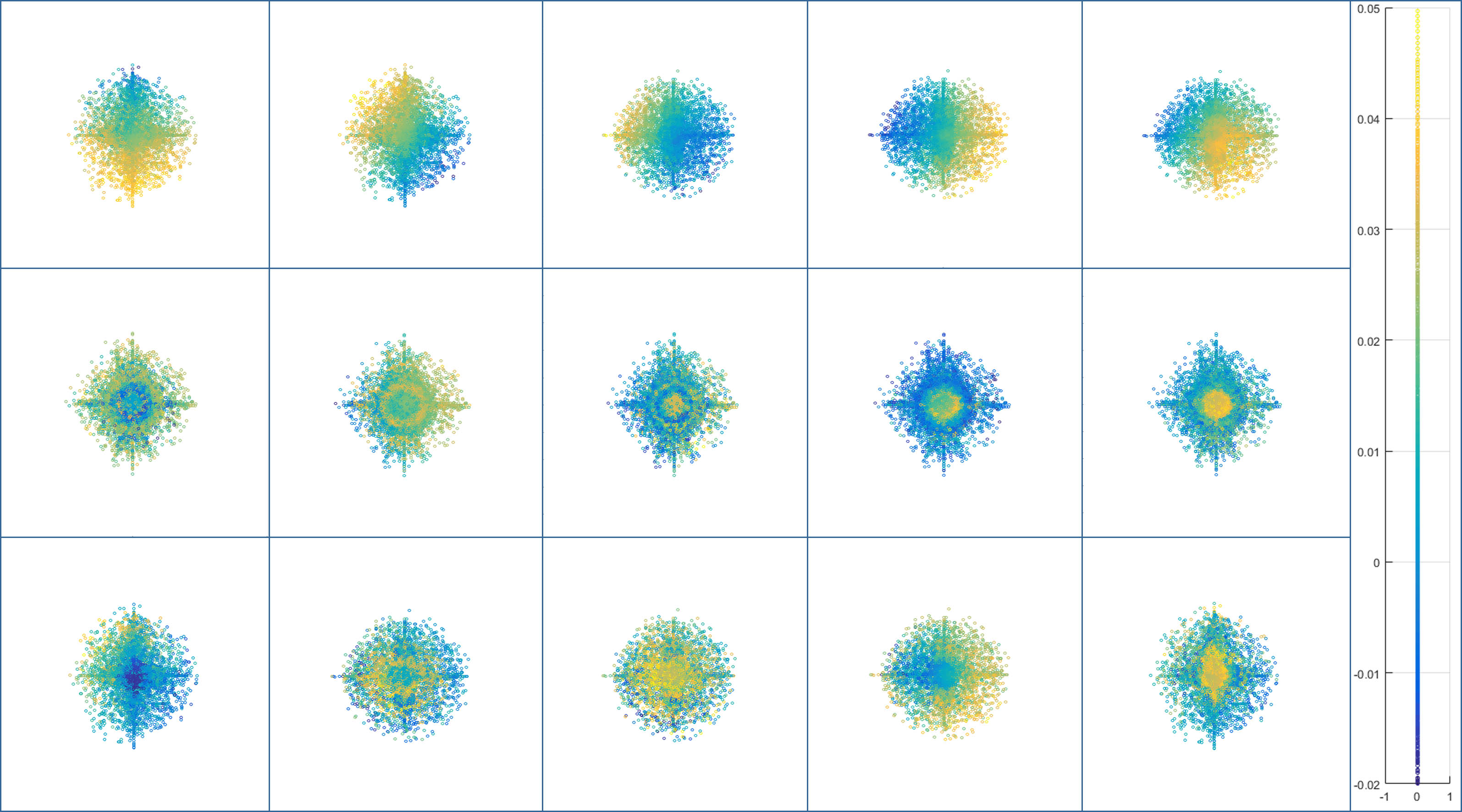} 
\caption{
Visualization for the convolutional filters learned in in the first layer of SpiderCNN. The 3D filters are shown as scatter plots projected on to the planes $x = 0$ or $y = 0$ or $z = 0$. 
}
\label{fig:filters2d}
\end{figure}

In Figure~\ref{fig:filters2d} we choose a plane passing through the origin, and project the points that lie on one side of the plane of the scatter graph onto the plane. We see some similar patterns that appear in 2D image filters. The visualization gives some hints about the geometric features that the convolutional filters in SpiderCNN learn. For example, the first row in  Figure~\ref{fig:filters2d} corresponds to 2D image filters that can capture boundary information.

\section{Conclusions}

A new convolutional neural network SpiderCNN that can directly process 3D point clouds with parameterized convolutional filters is proposed. More complex network architectures and more applications of SpiderCNN can be explored.

\section*{Acknowledgement}
This work was  supported by Shenzhen Basic Research Program (JCYJ201509251 63005055,  JCYJ20170818164704758), National Natural Science Foundation of
China (U1613211, 61633021, 61502263) and External Cooperation Program of BIC Chinese Academy of Sciences (172644KYSB20150019). We would like to thank Zhikai Dong for his technical assistance and helpful discussion. 

\section{Appendix}
In this section, we provide some additional details and experimental results. 

\subsection{Taylor v.s. MLP}

Recall that the filter used in SpiderCNN decomposes as 
\[
g(x, y, z) = g^{Step}(x, y, z) \cdot g^{Taylor}(x, y, z).
\]
We study empirically the effect of replacing $g^{Taylor}$ with MLP in a 4-layer SpiderCNN for classification on ModelNet40. The results are summarized in Table~\ref{table:mlp}. We see that Taylor outperformances MLP with more parameters.

\setlength{\tabcolsep}{4pt}
\begin{table}
\begin{center}
\caption{Taylor vs. MLP for classification on ModelNet40 }
\label{table:mlp}
\begin{tabular}{l|cccc}
\hline\noalign{\smallskip}
Methods & MLP(10, 5, 3) & MLP(6, 5, 3) & MLP(5, 3, 3) & order-3 Taylor\\
\noalign{\smallskip}
\hline
\noalign{\smallskip}
 Accuracy(\%) & 91.6 & 91.1 & 91.4 & 92.4
\\
\noalign{\smallskip}
\hline
\end{tabular}
\end{center}
\end{table}
\setlength{\tabcolsep}{1.4pt}

\subsection{Number of parameters in Taylor}

Recall that $g^{Taylor}$ is chosen to be order-3 expansion in SpiderCNN, and the trilinear interpolation gives us a simpler expansion:
\[
f_{w^T} = w^T_0 + w^T_1 x + w^T_2 y + w^T_3 z + w^T_4 xy + w^T_5 yz + w^T_6 xz + w^T_{16} xyz.
\]
We study the effect of replacing $g^{Taylor}$ with $f_{w^T}$, order-2 Taylor and linear Taylor in 4-layer SpiderCNN for classification on ModelNet40. The results are summarized in Table~\ref{table:Taylor}.

\setlength{\tabcolsep}{4pt}
\begin{table}
\begin{center}
\caption{The effect of number of parameters used in Taylor expansions on ModelNet40 classification accuracy.}
\label{table:Taylor}
\begin{tabular}{l|cccc}
\hline\noalign{\smallskip}
Method & order-3 Taylor & order-2 Taylor & linear Taylor & $f_{w^T}$
\\
\noalign{\smallskip}
\hline
\noalign{\smallskip}
Accuracy(\%) & 92.4 & 91.9 & 91.6 & 91.9
\\
\noalign{\smallskip}
\hline
\end{tabular}
\end{center}
\end{table}
\setlength{\tabcolsep}{1.4pt}

One justification for using oder-3 Taylor instead of $f_{w^T}$ is that if we compose $f_{w^T}$ with a rigid transformation $\mathbb{R}^3$, terms like $x^2y$ will appear. For instance, under the transformation $x' = x +y$, the expression $x'^2$ becomes $x^2 + 2xy + y^2$.

\subsection{SpiderCNN + PointNet}

\begin{figure}
\centering
\includegraphics[height=3cm]{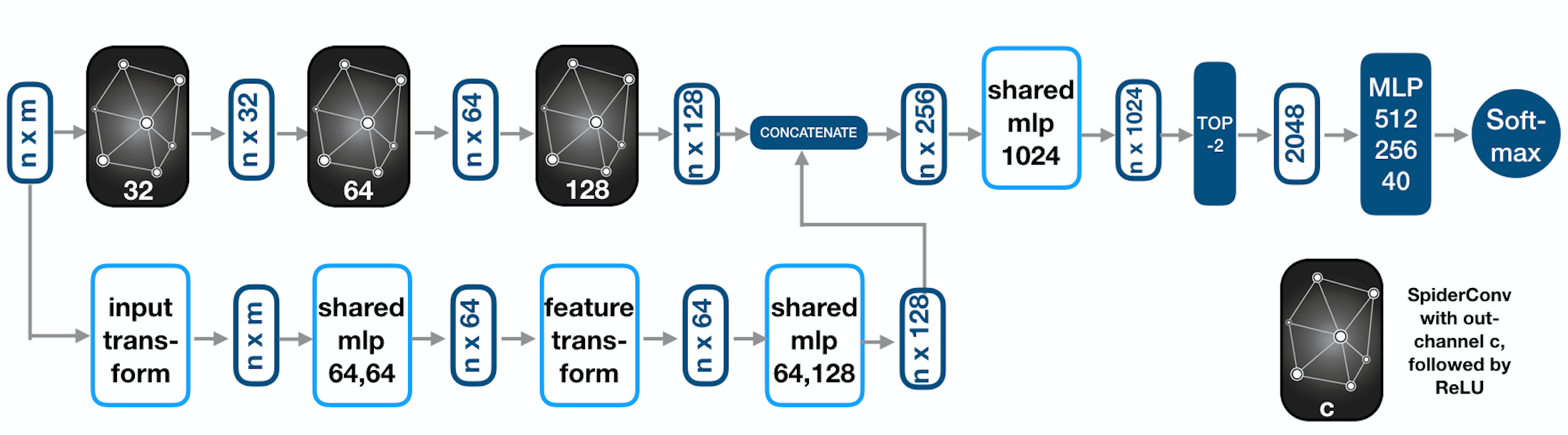} 
\caption{
An architecture that combines 3-layer SpiderCNN and PointNet.
}
\label{fig:pointnet}
\end{figure}

Figure~\ref{fig:pointnet} shows the architecture of combing a 3-layer SpiderCNN with PointNet in Section 4.1. Table~\ref{table:combine} summarizes the classification results on ModelNet40. We see that the combined model outperforms 3-layer SpiderCNN and PointNet.

\setlength{\tabcolsep}{4pt}
\begin{table}
\begin{center}
\caption{Classification results on ModelNet40 }
\label{table:combine}
\begin{tabular}{l|ccc}
\hline\noalign{\smallskip}
Methods & 3-layer SpiderCNN & PointNet & SpiderCNN + PointNet\\
\noalign{\smallskip}
\hline
\noalign{\smallskip}
 Accuracy(\%) & 91.5 & 90.3 & 92.2
\\
\noalign{\smallskip}
\hline
\end{tabular}
\end{center}
\end{table}
\setlength{\tabcolsep}{1.4pt}

%
%
%
\bibliographystyle{splncs04}
\bibliography{egbib}
%

\end{document}